\documentclass[10pt,twocolumn,letterpaper]{article}

\usepackage{cvpr}              % To produce the CAMERA-READY version
\usepackage{multirow}
\usepackage{array}
\usepackage{makecell}
\newcolumntype{P}[1]{>{\centering\arraybackslash}p{#1}}

\usepackage[dvipsnames]{xcolor}

\usepackage{indentfirst}

\definecolor{cvprblue}{rgb}{0.21,0.49,0.74}
\usepackage[pagebackref,breaklinks,colorlinks,citecolor=cvprblue]{hyperref}

\def\paperID{14418} % *** Enter the Paper ID here
\def\confName{CVPR}
\def\confYear{2024}

\title{Robust Multimodal 3D Object Detection \\ via Modality-Agnostic Decoding and Proximity-based Modality Ensemble}
\newcommand*\samethanks[1][\value{footnote}]{\footnotemark[#1]}
\author{\textbf{Juhan Cha}\thanks{equal contribution.} \hspace{0.4cm} \textbf{Minseok Joo}\samethanks \hspace{0.4cm} \textbf{Jihwan Park} \hspace{0.4cm} \textbf{Sanghyeok Lee} \hspace{0.4cm} \textbf{Injae Kim} \hspace{0.4cm} \textbf{Hyunwoo J. Kim}\thanks{Corresponding author.} \vspace{0.4cm} \\
Department of Computer Science and Engineering, Korea University\\
{\tt\small \{\href{mailto:hanchaa@korea.ac.kr}{hanchaa}, \href{mailto:wlgkcjf87@korea.ac.kr}{wlgkcjf87}, \href{mailto:jseven7071@korea.ac.kr}{jseven7071}, \href{mailto:cat0626@korea.ac.kr}{cat0626}, \href{mailto:dna9041@korea.ac.kr}{dna9041},
\href{mailto:hyunwoojkim@korea.ac.kr}{hyunwoojkim}\}@korea.ac.kr}
}

\begin{document}
\maketitle
\begin{abstract}
Recent advancements in 3D object detection have benefited from multi-modal information from the multi-view cameras and LiDAR sensors.
However, the inherent disparities between the modalities pose substantial challenges.
We observe that existing multi-modal 3D object detection methods heavily rely on the LiDAR sensor, treating the camera as an auxiliary modality for augmenting semantic details.
This often leads to not only underutilization of camera data but also significant performance degradation in scenarios where LiDAR data is unavailable.
Additionally, existing fusion methods overlook the detrimental impact of sensor noise induced by environmental changes, on detection performance.
In this paper, we propose \textbf{MEFormer} to address the LiDAR over-reliance problem by harnessing critical information for 3D object detection from every available modality while concurrently safeguarding against corrupted signals during the fusion process.
Specifically, we introduce Modality Agnostic Decoding (MOAD) that extracts geometric and semantic features with a shared transformer decoder regardless of input modalities and provides promising improvement with a single modality as well as multi-modality.
Additionally, our Proximity-based Modality Ensemble (PME) module adaptively utilizes the strengths of each modality depending on the environment while mitigating the effects of a noisy sensor.
Our MEFormer achieves state-of-the-art performance of 73.9\% NDS and 71.5\% mAP in the nuScenes validation set.
Extensive analyses validate that our MEFormer improves robustness against challenging conditions such as sensor malfunctions or environmental changes.
The source code is available at~\url{https://github.com/hanchaa/MEFormer}

\end{abstract}

\section{Introduction}
\label{sec:intro}

Multi-sensor fusion, which utilizes information from diverse sensors such as LiDAR and multi-view cameras, has recently become mainstream in 3D object detection~\cite{yan2023cmt, xie2023sparsefusion, liu2023bevfusion, Liang2022bevfusion, yang2022deepinteraction}.
Point clouds from the LiDAR sensor provide accurate geometric information of the 3D space and images from the multi-view camera sensors provide rich semantic information.
The effective fusion of these two modalities leads to state-of-the-art performance in 3D object detection by compensating for insufficient information of each modality.

\begin{figure*}[t]
    \centering
    \includegraphics[width=1\linewidth]{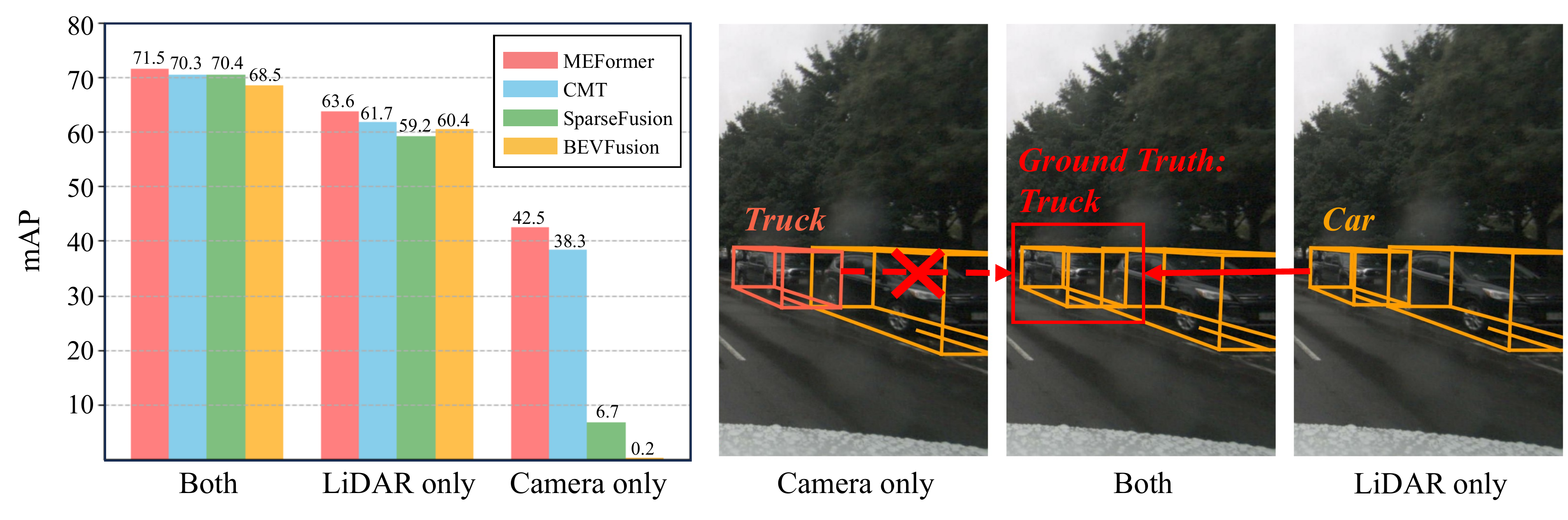}
    \vspace{-20pt}
    \caption{
        \textbf{Left}: Comparison of performance drop in sensor missing scenarios.
        MEFormer shows the smallest performance degradation compared to previous works.
        Specifically, CMT~\cite{yan2023cmt} shows 32\% mAP drop and BEVFusion~\cite{liu2023bevfusion} shows 68.3\% mAP drop while ours shows only 29\% mAP drop in camera only scenario.
        \textbf{Right}: Illustration of negative fusion.
        Although the prediction by a single modality (\eg, Camera) is correct,  
        the multimodal predictions are often negatively affected by inaccurate unimodal signals, resulting in misclassification.
    }
    \label{fig:fig1}
    \vspace{-12pt}
\end{figure*}

However, as discussed in Yu et al.~\cite{yu2023benchmark}, previous frameworks~\cite{Bai2022transfusion, wang2021pointaugmenting, li2022deepfusion, vora2020pointpainting} have LiDAR reliance problem that primarily relies on LiDAR modality and treats camera modality as an extra modality for improving semantic information, even though recent studies~\cite{liu2022petr, li23bevdepth, li2022bevformer} show that the geometric information can also be extracted solely from camera modality.
The LiDAR reliance problem makes the framework fail to extract geometric information from camera modality, which results in missing objects that can only be found by the camera~\eg, distant objects.
This problem is exacerbated when the LiDAR sensor is malfunctioning.
In LiDAR missing scenarios during inference, although the model is trained with both modalities, it shows inferior performance to the same architecture trained only with camera modality or even completely fails to perform the detection task (see the left graph of~\cref{fig:fig1}).
Moreover, previous works~\cite{liu2023bevfusion, Liang2022bevfusion} simply fuse the point feature and image pixel feature in the same coordinate in Bird's-Eyes-View (BEV) space without considering the disparities between two modalities.
In a challenging environment where one modality exhibits weak signals, such as at night time, prior works may suffer from a negative fusion that the defective information from a noisy modality adversely affects the correct information obtained from another modality, which results in corrupted detection performance as shown in the right side of~\cref{fig:fig1}.

In this paper, we introduce the \textbf{M}odality \textbf{E}nsemble trans\textbf{Former}, dubbed MEFormer, which effectively leverages the inherent characteristics of both LiDAR and camera modalities to tackle the LiDAR reliance and negative fusion problems.
First, inspired by multi-task learning~\cite{nekrasov2019mtl, zhang2022mtl_survey}, we propose the \textbf{Mo}dality-\textbf{A}gnostic \textbf{D}ecoding (MOAD) training strategy to enhance the ability of the transformer decoder in extracting both geometric and semantic information regardless of input modalities.
In addition to the multi-modal decoding branch, which takes both modalities as input, we introduce auxiliary tasks, where the transformer decoder finds objects with single-modal decoding branches that take only a single modality as input.
This alleviates the LiDAR reliance problem by enabling the decoder to extract the information needed for 3D object detection from individual modalities.
In addition, we propose \textbf{P}roximity-based \textbf{M}odality \textbf{E}nsemble (PME), which alleviates the negative fusion problem.
A simple cross-attention mechanism with our proposed attention bias generates a final box prediction by integrating the box candidates from all modality decoding branches.
PME adaptively aggregates the box features from each modality decoding branch depending on the environment and mitigates the noise that may occur in the multi-modal decoding branch.
Extensive experiments demonstrate that MEFormer exhibits superior performance.
Especially in challenging environments such as sensor malfunction, MEFormer shows robust performance compared to previous works.

Our contributions are summarized as:
\begin{itemize}
    \item We propose a novel training strategy Modality-agnostic Decoding (MOAD), which is effective in addressing the LiDAR reliance problem by better utilizing information from all modalities.
    \item We introduce the Proximity-based Modality Ensemble (PME) module, which adaptively aggregates box predictions from three decoding branches of MOAD to prevent the negative fusion problem.
    \item MEFormer achieves the state-of-the-art 3D object detection performance on nuScenes dataset. We also show promising performance in challenging environments.
\end{itemize}
\vspace{-4pt}
\section{Related Work}
\label{sec:rel}
\begin{figure*}[t!]
    \centering
    \includegraphics[width=0.98\textwidth]{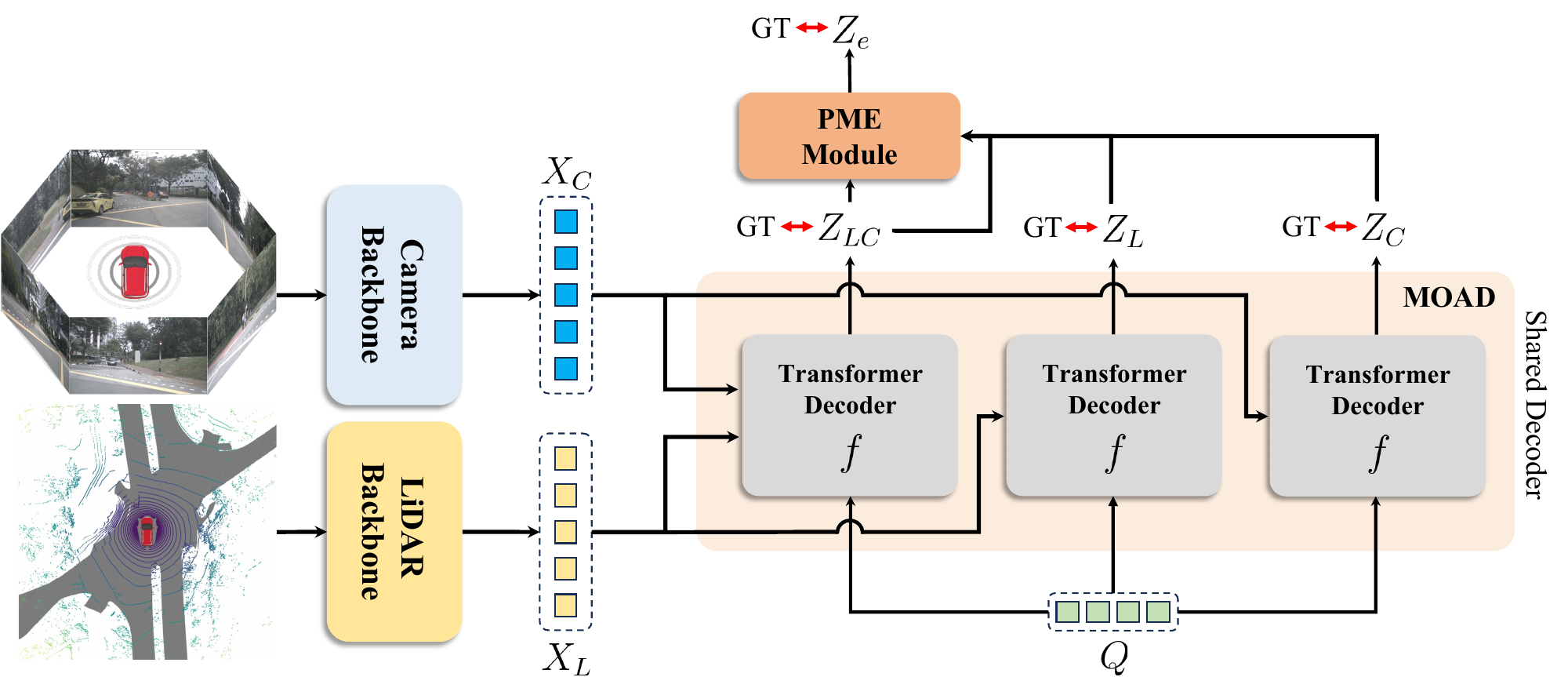}
    \caption{
        \textbf{The overall architecture of MEFormer.}
        In our framework, we employ two modalities: one dedicated to the image (camera) and the other to the point clouds (LiDAR).
        The camera and LiDAR backbones simultaneously extract the feature maps from both the image and point clouds.
        Then, three modality decoding branches process the initial object query $Q$.
        Each modality decoding branch has the transformer decoder $f$ which shares parameters with all other modality decoding branches.
        Each uses different combinations of modalities as key and value, \eg, LiDAR + camera (LC), LiDAR (L), and camera (C), resulting in the box features $Z_{LC}$, $Z_{L}$, and $Z_{C}$, respectively.
        During training, the predicted boxes of each modality decoding branch are separately supervised by ground truth boxes.
        Finally, the PME module based on cross-attention acquires $Z_{LC}$ for query and $Z_{LC}, Z_L$, and $Z_C$ for key and value, generating the final ensembled box features $Z_e$.
        The predicted boxes derived from $Z_e$ are also supervised by the ground truth boxes.
    }
    \label{fig:main}
    \vspace{-13pt}
\end{figure*}
\noindent\textbf{Camera-based 3D Object Detection.}
The field of camera-based 3D object detection has witnessed numerous recent advances in autonomous driving. 
Some previous studies~\cite{philion2020lift, huang2021bevdet, reading2021categorical, wang2019pseudo}, have proposed a method to lift 2D features into 3D space by predicting pixel-wise depth from a camera.
Despite its simplicity and good performance, this approach is constrained by its dependence on accurate depth prediction.
Other works~\cite{li2022bevformer, yang2023bevformerv2, liu2022petr, liu2023petrv2} introduce leveraging 3D queries and employing transformer attention mechanisms to find the corresponding 2D features. In particular, PETR~\cite{liu2022petr} showed efficacy in encoding 3D coordinates into image features through positional encoding, implicitly deducing 3D-2D correspondences and achieving good performance.
However, relying solely on a camera sensor for 3D perception, while advantageous due to its lower cost than a LiDAR sensor, faces limitations stemming from the inherent ambiguity in predicting 3D features from the 2D image.

\noindent\textbf{Multi-modal 3D Object Detection.}
In the domain of 3D object detection for autonomous driving, multi-modal detection methods that leverage data from both LiDAR sensors and multi-view cameras achieve state-of-the-art performance. 
These two sensors provide complementary information to each other, prompting numerous studies~\cite{qi2018frustum, yin2021mvp, vora2020pointpainting, wang2021pointaugmenting, li2022uvtr, chen2023futr3d, chen2022autoalignv2} to explore methodologies for learning through these both modality.
TransFusion~\cite{Bai2022transfusion} and DeepFusion~\cite{li2022deepfusion} introduce a transformer-based method that utilizes LiADR features as queries, and image features as keys and values. 
Meanwhile, BEVFusion~\cite{liu2023bevfusion,Liang2022bevfusion} shows commendable performance by lifting 2D features into a unified Bird's Eye View (BEV) space following LSS~\cite{philion2020lift}, and subsequently applying a 3D detector head.
DeepInteraction~\cite{yang2022deepinteraction} applies iterative cross-attention by extracting queries from each modality to fully exploit modality-specific information. 
SparseFusion~\cite{xie2023sparsefusion} learns sparse candidates acquired via modality-specific detectors and fuses these sparse candidates to generate the final outputs.
CMT~\cite{yan2023cmt} aggregates information into a query with the assistance of modality positional embedding to generate final boxes. However, most previous studies are highly dependent on LiDAR modality, not fully exploiting the information inherent in each modality. 
Furthermore, the modality fusion methods often overlook the distinctions between modalities, leading to a degradation of robustness in scenarios where a specific sensor introduces noise.
% In this paper, we make full use of important information from each modality and propose an effective method that is robust even in challenging environments.

\noindent\textbf{Robust Multi-Modality Fusion.}
In real-world driving scenarios, sensor failures are prevalent and adversely affect the stability required for autonomous driving~\cite{xie2023robobev, ge2023metabev, zhang2019robust, yu2023benchmark, song2024robustness, zhu2023understanding}.
Apart from a simple sensor missing, external environment noise or sensor malfunctions can severely impair robust 3D perception.
While many studies~\cite{Bai2022transfusion, liu2023bevfusion, Liang2022bevfusion, li2022deepfusion, yang2022deepinteraction, xie2023sparsefusion} have explored robust 3D detection through multi-sensor fusion, they primarily focus on achieving superior performance on complete multi-modal inputs. 
However, reliance on an ideal sensor results in significant performance degradation when a specific sensor malfunctions, diminishing the effectiveness of the fusion approach compared to utilizing a single modality.
Specifically, existing fusion methods heavily rely on LiDAR modality, and the camera is used in an auxiliary role, resulting in inferior performance in the case of LiDAR corruption~\cite{yu2023benchmark}.
% % While previous studies have explored robust 3D detection through multi-sensor fusion, they primarily focus on input modality augmentation, resulting in limited robustness due to a lack of consideration for modality-specific characteristics. 
% % Notably, existing multi-modal fusion methods tend to heavily rely on LiDAR, leading to significant performance degradation in the case of LiDAR failure. 
% Notably, current fusion methods do not take advantage of fusion because the image modality cannot effectively mitigate LiDAR noise.
In this paper, we propose a framework that reduces reliance on specific modalities, mitigates the negative fusion problem, and achieves robust detection performance even in scenarios involving noise or sensor missing.
\vspace{-4pt}
\section{Method}
\label{sec:method}
In this section, we propose MEFormer, a robust 3D object detection framework with modality-agnostic decoding and modality ensemble module.
The overall architecture is illustrated in~\cref{fig:main}.
We start with a brief review of the cross-modal transformer in~\cref{method:transformer}, and then, in~\cref{method:moad}, we present the modality-agnostic decoding that enables fully extracting features for 3D object detection from each modality to reduce reliance on LiDAR modality.
Finally, in~\cref{method:ensemble}, we introduce a proximity-based modality ensemble module to adaptively integrate box predictions from each decoding branch while preventing negative fusion in the modality fusion process.
    \subsection{Preliminaries}
\label{method:transformer}
Cross modal transformer (CMT)~\cite{yan2023cmt} is a recent framework that uses a transformer decoder to aggregate information from both modalities into object queries.
In CMT, the modality-specific backbone first extracts modality features~\eg, VoVNet~\cite{lee2020vovnet} for camera and VoxelNet~\cite{zhou2018voxelnet} for LiDAR.
Then, they localize 3D bounding boxes using a transformer decoder $f$ with modality-specific position embeddings that help object queries aggregate information from both modalities simultaneously.
Given the flattened LiDAR BEV features $X_L\in\mathbb{R}^{(H_L*W_L)\times D}$ and camera image features $X_C\in\mathbb{R}^{(N_C*H_C*W_C)\times D}$, CMT can be formulated as: 
\begin{equation}
    Z=f(Q, [X_L; X_C]),
\end{equation}
where $Q$ denotes a set of learnable object queries and [;] indicates the concatenation.
$H_L, W_L, H_C,$ and $W_C$ denote the height and width of the LiDAR BEV feature map and camera image feature map respectively, and $N_C$ indicates the number of cameras.

CMT is an effective framework but it has some drawbacks discussed in~\cref{sec:intro}.
First, when a LiDAR is missing, CMT lacks the ability to extract geometric information from the camera modality, resulting in substantial performance degradation.
Second, when a specific modality shows corrupted signals, information from it may act as noise when aggregating information from both modalities simultaneously.
In the following sections, we will demonstrate how to mitigate these issues.
    \subsection{Modality-Agnostic Decoding}
\label{method:moad}
For the cross-modal transformer to maximize the use of both modalities, both geometric and semantic information should be extracted from each modality without relying on a specific one.
To this end, we propose a modality-agnostic decoding (MOAD) training scheme, that allows a transformer decoder to fully decode information for 3D object detection regardless of input modalities.

First, we randomly sample learnable anchor positions in the 3D space from uniform distributions and set their positional embeddings to the initial object queries $Q\in\mathbb{R}^{N\times D}$ following PETR~\cite{liu2022petr}.
Then $Q$ is processed by multiple decoding branches, each using different modality combinations as an input.
This can be formulated as:
\begin{align}
    &Z_{LC}=f(Q, [X_L;X_C]),\\
    &Z_L=f(Q, X_L),\\
    &Z_C=f(Q, X_C),
\end{align}
where $f$ is a shared transformer decoder and $Z_{LC}, Z_L$, and $Z_C$ denote box features from each modality decoding branch.
Note that $Q$ is shared across multiple decoding branches.
We generate modality-specific positional embeddings $\Gamma_L$ and $\Gamma_C$ following~\cite{liu2022petr, yan2023cmt}, and $\Gamma_L + \Gamma_C, \Gamma_L$, and $\Gamma_C$ are used as positional embedings for queries when generating $Z_{LC}, Z_L$, and $Z_C$ respectively.
Then, we predict the final box prediction $\hat{B}_m$ and classification score $\hat{P}_m$ via modality-agnostic box head $h$:
\begin{align}
    &\hat{B}_m=h^{\text{reg}}(Z_m),\ \hat{P}_m=h^{\text{cls}}(Z_m),
\end{align}
where $m=\{LC, L, C\}$ indicates the set of modality decoding branch.

We use Hungarian matching between ground truth boxes and predicted boxes from each modality decoding branch respectively for loss computation.
The loss function from each modality decoding branch $\mathcal{L}_m$ can be formulated as:
\begin{align}
\label{eq:loss}
    \mathcal{L}_m=\omega_{\text{reg}}\mathcal{L}_{\text{reg}}(B, \hat{B}_m) + \omega_{\text{cls}}\mathcal{L}_{\text{cls}}(P, \hat{P}_m),
\end{align}
and overall loss function of MOAD is defined as:
\begin{align}
    \mathcal{L}_{\text{MOAD}}=\omega_{LC}\mathcal{L}_{LC}+\omega_L\mathcal{L}_L+\omega_C\mathcal{L}_C.
\end{align}
We use the focal loss for classification loss and L1 loss for box regression.

Note that applying Hungarian matching and computing loss function separately for each modality decoding branch helps fully decode information regardless of input modality.
By minimizing the shared transformer decoder with the loss function $\mathcal{L}_L$ and $\mathcal{L}_C$, modality backbones and transformer decoder learn to extract both geometric and semantic information from each modality without relying o3n specific modality.
In addition, minimizing $\mathcal{L}_{LC}$ helps the shared transformer learn how to fuse rich information from both modalities effectively.
At test time, we only use the multi-modal decoding branch for final box prediction, resulting in higher performance without additional computational cost.
\vspace{-8pt}
    \begin{figure}[t]
    \centering
    \includegraphics[width=0.95\linewidth]{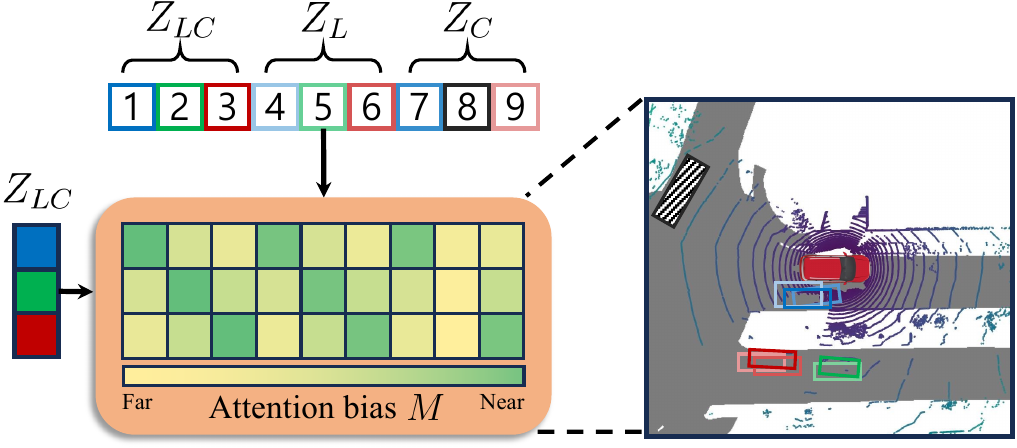}
    \caption{
    \textbf{Illustration of our proximity-based modality ensemble module.}
    PME takes box features $Z_{LC}$ as query and $Z_{LC}, Z_L$, and $Z_C$ as key and value.
    To reduce the interaction between irrelevant boxes, we calculate the attention bias $M$ based on the center distance between the predicted boxes.
    Then, we add attention bias $M$ to the attention logit before applying the softmax function.
    }
    \label{fig:concept}
    \vspace{-14pt}
\end{figure}
\subsection{Proximity-based Modality Ensemble}
\label{method:ensemble}
As discussed in~\cite{wang2023challenge, mao2023survey}, there are challenging environments where one modality outperforms the other, such as at night or sensor malfunction environments.
In this case, aggregating information from both LiDAR and camera modalities simultaneously may suffer from negative fusion, which leads to poor box prediction as discussed in~\cref{sec:intro}.
To this end, we propose a novel proximity-based modality ensemble (PME) module that adaptively integrates box features from every modality decoding branch of MOAD.

Given the set of the box features $Z_m$ from each modality decoding branch, $Z_{LC}$ is updated through a cross-attention layer with the whole output features including itself.
To avoid the noise transfer caused by the interaction between irrelevant box features, we introduce an attention bias $M$ for the cross-attention.
To obtain the attention bias, we measure the center distance between the predicted boxes from $Z_{LC}$ and those from $Z_{LC}, Z_L$, and $Z_C$.
Then, we apply a linear transformation with a learnable scaler $\alpha\in\mathbb{R}^1$ and a bias $\beta\in\mathbb{R}^1$.
Given $Z_m$ and their corresponding center $x, y$ coordinates $\hat{\Phi}_m=\{\hat{\phi}_{m, 1}, \ldots, \hat{\phi}_{m, N}\}$ in the BEV space, the attention bias $M\in\mathbb{R}^{N\times3N}$ can be formulated as:
\begin{align}
    &\hat{\Phi}_A=[\hat{\Phi}_{LC};\hat{\Phi}_L;\hat{\Phi}_C],\\
    &M_{i, j} = \alpha * ||\hat{\phi}_{LC, i} - \hat{\phi}_{A, j}|| + \beta.
    % &M_{i, j} = \left\{
    %     \begin{array}{ll}
    %     \label{eq:mask}
    %         1, & \text{if}\ ||\hat{\phi}_{LC,i} - \hat{\phi}_{A, j}|| \le d,\\
    %         0, & \text{otherwise},
    %     \end{array}
    % \right.
\end{align}
Then, we add attention bias $M$ to the attention logit and apply the softmax function to calculate attention scores.
% We select $d$ from distance threshold candidates of nuScenes~\cite{caesar2020nuscenes} mAP metric and empirically set it to 2 meters.

Every set of the box features $Z_{LC}, Z_L$, and $Z_C$ are linearly projected by modality-specific projection function $g_{LC}, g_L$, and $g_C$ before the cross-attention layer.
To summarize, the overall architecture of the PME can be written as:
\begin{align}
    &Q_m'=g_m(Z_m),\ \forall m\in\{LC, L, C\},\\
    &Z_e=f_e(Q_{LC}', [Q_{LC}';Q_L';Q_C'], M),
\end{align}
where $f_e$ is a cross-attention layer for the modality ensemble.
We generate positional embeddings for queries and keys using $\hat{\Phi}_{LC}, \hat{\Phi}_L$, and $\hat{\Phi}_C$ and MLPs.
The final box prediction $\hat{B}_e$ and $\hat{P}_e$ are generated by the box head $h_e$ which is another box head for the ensembled box features $Z_e$, as:
\begin{align}
    \hat{B}_e=h_e^{\text{reg}}(Z_e),\ \hat{P}_e=h_e^{\text{cls}}(Z_e).
\end{align}

For proximity-based modality ensemble loss, we apply Hungarian matching between ground truth boxes and predicted boxes.
The loss function of PME is defined as:
\begin{align}
    \mathcal{L}_{\text{PME}}=\omega_{\text{reg}}\mathcal{L}_{\text{reg}}(B,\hat{B}_e) + \omega_{\text{cls}}\mathcal{L}_{\text{cls}}(P, \hat{P}_e),
\end{align}
where $\mathcal{L}_{\text{reg}}$ and $\mathcal{L}_{\text{cls}}$ is the L1 loss and the focal loss respectively following~\cref{eq:loss}.

Note that the cross-attention mechanism helps our framework adaptively select promising modalities depending on the environment when predicting the final box.
MEFormer with PME shows remarkable performance in challenging environments, which will be discussed in~\cref{exp:robustness}.
\section{Experiments}
\label{sec:experiments}
In this section, we present comprehensive experiments of our framework. First, we introduce the dataset, metrics, and experimental settings. Then we demonstrate the effectiveness of MEFormer through comparison with state-of-the-art methods on the benchmark dataset. Additionally, we analyze the contribution of the proposed framework through ablations and extensive experiments about various scenarios.
    \subsection{Experimental Setup}
\label{exp:setup}
\noindent\textbf{Datasets and Metrics.}
We evaluate MEFormer on the nuScenes dataset~\cite{caesar2020nuscenes}, a large-scale benchmark dataset for 3D object detection. 
It includes point clouds collected with 32-beam LiDAR and 6 multi-view images of 1600$\times$900 resolution. 
The dataset is composed of 1000 scenes and split into 700, 150, and 150 scenes for training, validation, and testing.
We report 3D object detection performance through mAP and NuScenes Detection Score (NDS). 
Unlike the conventional Average Precision metric, nuScenes mAP determines the box matches by considering center distance in Bird's Eye View (BEV) space.
A prediction is considered positive if the ground truth box lies within a certain distance from the center of the prediction.
This metric averages results across 10 classes with 4 distance thresholds (0.5m, 1m, 2m, 4m).
NDS represents an integrated metric considering mAP and various true positive metrics, which consist of translation, scale, orientation, velocity, and attributes.

    \subsection{Implementation details}
\label{exp:detail}
\noindent\textbf{Model.}
We implement MEFormer with MMDetection3D framework~\cite{mmdet3d2020}.
We use VoVNet~\cite{lee2020vovnet} for the camera modality backbone and use images with a resolution of 1600 $\times$ 640 by cropping the upper region of the image.
For the LiDAR modality, VoxelNet~\cite{zhou2018voxelnet} is used as a backbone.
We set the detection range to [$-54.0m, 54.0m$] and [$-5.0m, 3.0m]$ for the XY-axes and Z-axis, respectively, and the voxel size is set to $0.075m$.
Our shared transformer decoder $f$ of MOAD has 6 cross-attention layers and PME has 1 cross-attention layer.

\begin{table*}[t!]
\centering
\setlength{\tabcolsep}{15pt}
\begin{tabular}{lccccc}
\toprule
\multirow{2}{*}{Method} & \multirow{2}{*}{Modality}  & \multicolumn{2}{c}{Validation set} & \multicolumn{2}{c}{Test set}\\
\cmidrule(lr){3-4} \cmidrule(lr){5-6}
 &  & {NDS} & {mAP} &{NDS} &{mAP}\\
\midrule
\midrule
{BEVDet~\cite{huang2021bevdet}} & {C} & {-} & {-} & {48.2} & {42.2} \\
{FCOS3D~\cite{wang2021fcos3d}} & {C} & {41.5} & {34.3} & {42.8} & {35.8} \\
{PETR~\cite{liu2022petr}} & {C} & {44.2} & {37.0} & {45.5} & {39.1} \\
\midrule
{SECOND~\cite{yan2018second}}      & {L}  & {63.0} & {52.6} & {63.3} & {52.8} \\
{CenterPoint~\cite{yin2021center}} & {L}  & {66.8} & {59.6} & {67.3} & {60.3} \\
{TransFusion-L~\cite{Bai2022transfusion}} & {L}  & {70.1} & {65.1} & {70.2} & {65.5} \\
\midrule
{PointAugmenting~\cite{wang2021pointaugmenting}} & {L+C}  & {-} & {-} & {71.0} & {66.8} \\
{FUTR3D~\cite{chen2023futr3d}} & {L+C}  & {68.3} & {64.5} & {-} & {-} \\
{UVTR~\cite{li2022uvtr}} & {L+C}  & {70.2} & {65.4} & {71.1} & {67.1} \\
{AutoAlignV2~\cite{chen2022autoalignv2}} & {L+C}  & {71.2} & {67.1} & {72.4} & {68.4} \\
{TransFusion~\cite{Bai2022transfusion}}       & {L+C}  & {71.3} & {67.5} & {71.6} & {68.9} \\
{BEVFusion~\cite{Liang2022bevfusion}}         & {L+C}  & {72.1} & {69.6} & {73.3} & {71.3} \\
{BEVFusion~\cite{liu2023bevfusion}}         & {L+C}  & {71.4} & {68.5} & {72.9} & {70.2} \\
{DeepInteraction~\cite{yang2022deepinteraction}}   & {L+C}  & {72.6} & {69.9} & {73.4} & {70.8} \\
{UniTR~\cite{wang2023unitr}}             & {L+C}  & {\underline{73.3}} & {\underline{70.5}} & {\textbf{74.5}} & {70.9} \\
{MetaBEV~\cite{ge2023metabev}}             & {L+C}  & {71.5} & {68.0} & {-} & {-} \\
{SparseFusion~\cite{xie2023sparsefusion}}      & {L+C}  & {72.8} & {70.4} & {73.8} & {\underline{72.0}} \\
{CMT~\cite{yan2023cmt}}               & {L+C}  & {72.9} & {70.3} & {74.1} & {\underline{72.0}} \\
\midrule
{\textbf{MEFormer (Ours)}} & {L+C}  & {\textbf{73.9}} & {\textbf{71.5}} & {\underline{74.3}} & {\textbf{72.2}} \\
\bottomrule
\end{tabular}
\caption{
\textbf{Results on the nuScenes validation and test set.}
}
\label{main_table}
\end{table*}
\noindent\textbf{Training.}
All experiments are conducted with a batch size of 16 on 8 A6000 GPUs.
Our model is trained through two steps: 1) We first train MOAD without PME for 20 epochs with CBGS~\cite{zhu2019cbgs}.
We apply GT sampling augmentation for the first 15 epochs and we do not apply for the last 5 epochs.
The initial learning rate is 1.0 $\times 10^{-4}$ and the cyclic learning rate policy~\cite{smith2017cyclic} is adapted.
2) Once the MOAD is trained, we train PME module for 6 epochs with CBGS while freezing the modality backbone and the rest of the transformer.
Note that we do not apply GT sampling augmentation in the second stage.
The initial learning rate is 1.0 $\times 10^{-4}$ and we adopt the cosine annealing learning rate policy with 1000 warm-up iterations~\cite{loshcilov2019cosine}.
AdamW~\cite{loshchilov2019adamw} optimizer is adopted for optimization in both stages.
For the loss weights, we set the $\omega_{\text{reg}}$ and $\omega_{\text{cls}}$ to 2.0 and 0.25 respectively following DETR3D~\cite{wang2021detr3d}.
We empirically set $\omega_{LC}, \omega_{L}$, and $\omega_{C}$ to 1.0 for MOAD.
    \subsection{Comparison to the state of the art framework}
\label{exp:comparison}
We compare our proposed model with existing baseline models on the nuScenes dataset. 
To evaluate the model on the test set, we submitted the detection results of MEFormer to the nuScenes test server and report the performance.
Note that we did not use any test-time augmentation or ensemble strategies during the inference. 
As shown in Table~\ref{main_table}, MEFormer outperforms other baseline models in terms of both mAP and NDS with a significant margin. Specifically, MEFormer attains a remarkable 73.9\% NDS and 71.5\% mAP on the nuScenes validation set. 
On the test set, our model achieves the best performance with 72.2\% mAP and the second best performance with 74.3\% NDS compared to the previous methods. 
This demonstrates that, although MEFormer is originally proposed for tackling the sensor missing or corruption scenarios, our proposed approach is effective for enhancing overall detection performance as well.
    \subsection{Robustness in challenging scenarios}
\label{exp:robustness}
        \begin{table*}[t!]
\centering
\setlength{\tabcolsep}{15pt}
\begin{tabular}{lccccccc}
    \toprule
    \multirow{2}{*}{Method} & \multirow{2}{*}{Modality} & \multicolumn{2}{c}{{Both}} & \multicolumn{2}{c}{{LiDAR only}} & \multicolumn{2}{c}{{Camera only}} \\
    \cmidrule(lr){3-4} \cmidrule(lr){5-6} \cmidrule(lr){7-8}
    &  & NDS & mAP & NDS & mAP & NDS & mAP \\
    \midrule
    \midrule
    {PETR~\cite{liu2022petr}} & C & -& -& -& -& 44.2& 37.0 \\
    {CMT-C~\cite{yan2023cmt}} & C & -& -& -& -& 46.0& 40.6 \\
    {CMT-L~\cite{yan2023cmt}} & L & -& -& 68.6& 62.4& -& - \\
    \midrule
    {BEVFusion~\cite{liu2023bevfusion}$\dagger$} & L+C & 71.4& 68.5& 66.5& 60.4& 1.3&0.2 \\
    {MetaBEV~\cite{ge2023metabev}} & L+C & 71.5& 68.0& 69.2& \textbf{63.6}& 42.6& 39.0\\
    {CMT~\cite{yan2023cmt}} & L+C &72.9& 70.3& 68.1& 61.7& 44.7& 38.3 \\
    \midrule
    {\textbf{Ours}} & L+C & \textbf{73.9}& \textbf{71.5}& \textbf{69.5} & \textbf{63.6} & \textbf{48.0} & \textbf{42.5} \\
    
    \bottomrule
\end{tabular}
\caption{
\textbf{Comparison of detection performance in sensor missing scenarios.}
Here, in the case where one sensor is missing, \textbf{Ours} does not apply the PME.
$\dagger$ denotes the results obtained using the OpenPCDet~\cite{openpcdet2020} reproduced weights.
}
\label{sensor_missing_table}
\end{table*}
\begin{table*}[t!]
\centering
\resizebox{\linewidth}{!}{
\setlength{\tabcolsep}{11pt}
\begin{tabular}{lc cc cc cc cc}
    \toprule
    \multirow{2}{*}{Method} & \multirow{2}{*}{Modality} & \multicolumn{2}{c}{\makecell{LiDAR\\Beam Reduction}} & \multicolumn{2}{c}{\makecell{Camera\\Dirt Occlusion}} & \multicolumn{2}{c}{Night} & \multicolumn{2}{c}{Rainy}\\
    \cmidrule(lr){3-4} \cmidrule(lr){5-6} \cmidrule(lr){7-8} \cmidrule(lr){9-10} & & NDS & mAP & NDS & mAP & NDS & mAP & NDS & mAP\\
    
    \midrule
    \midrule
    
    PETR~\cite{liu2022petr}$\ddagger$ & C & - & - & 30.6 & 17.9 & 24.2 & 17.2 & 50.6 & 41.9 \\
    Transfusion-L~\cite{Bai2022transfusion}$\dagger$ & L & 49.6 & 31.8 & - & - & 43.5 & 37.5 & 69.9 & 64.0 \\
    BEVFusion~\cite{liu2023bevfusion}$\dagger$ & L+C & 55.3 & 43.2 & 68.9 & 63.7 & 45.7 & 42.2 & 72.1 & 68.1 \\
    DeepInteraction~\cite{yang2022deepinteraction}$\ddagger$ & L+C & 55.1 & 46.0 & 65.9 & 63.8 & 43.8 & 42.3 & 70.6 & 69.4 \\
    \midrule
    CMT~\cite{yan2023cmt} & L+C & 62.2 & 54.9 & 69.2 & 63.9 & 46.3 & 42.8 & 73.7 & 70.5 \\
    \textbf{CMT w/ PME} & L+C & 62.4 & 55.1 & 69.5 & 64.5 & 46.3 & 43.0 & 74.0 & 70.7 \\
    \midrule
    \textbf{MOAD} & L+C & 62.8 & 55.0 & 69.7 & 64.3 & 46.6 & 43.1 & 74.6 & \textbf{72.2} \\
    \textbf{MOAD w/ PME} & L+C & \textbf{63.4} & \textbf{55.9} & \textbf{69.9} & \textbf{64.6} & \textbf{46.8} & \textbf{43.7} & \textbf{74.9} & \textbf{72.2} \\
    \bottomrule
\end{tabular}
}
\caption{
\textbf{Comparison of detection performance in challenging scenarios.}
For the LiDAR malfunction scenario, we apply beam reduction to 4 beams following BEVFusion~\cite{liu2023bevfusion}.
For the camera, we randomly overlap dirt masks onto the camera images following~\cite{yu2023benchmark}.
$\dagger$ and $\ddagger$ denote the results obtained using the OpenPCDet~\cite{openpcdet2020} reproduced weights and weight provided in the official GitHub repository respectively.
}
\label{sensor_corruption_table}
\vspace{-13pt}
\end{table*}
\vspace{-4pt}
\subsubsection{Sensor missing}
\vspace{-5pt}
We introduce modality-agnostic decoding, a novel training strategy for leveraging a shared decoder to extract both geometric and semantic information from each modality while reducing heavy reliance on specific modalities.
To validate this, Table~\ref{sensor_missing_table} presents the performance in environments where a single modality is missing at test time.
Under the absence of a camera or LiDAR sensor, our modality-agnostic decoding (w/o PME) shows strong robustness, demonstrating a significant performance gap compared to other baselines.
Specifically, in the camera-only scenario, CMT shows 28.2\% performance degradation in NDS while ours shows 25.9\%.
Note that BEVFusion is completely impaired if only cameras are available, which means BEVFusion fails to extract geometric information.
Additionally, the comparison between CMT-C, CMT-L, and CMT shows that when one modality is missing, the framework trained with both LiDAR and camera modalities exhibits a performance degradation compared to those trained with a single modality.
This means CMT does not fully utilize both geometric and semantic information from each modality, instead, it heavily relies on a specific modality.
However, our framework shows the best performance by 69.5\% NDS and 63.6\% mAP in the LiDAR-only scenarios and 48.0\% NDS and 42.5\% mAP in the camera-only scenarios.
Through the application of modality-agnostic decoding and the parameter-sharing mechanism in the transformer decoder, our framework demonstrates better robustness in both LiDAR-only and camera-only scenarios while mitigating the reliance on either modality.
        \vspace{-12pt}
\subsubsection{Challenging scenarios}
\label{exp:sensor_corruption}
\vspace{-6pt}
We propose modality-agnostic decoding to address the LiDAR reliance problem and effectively extract the information for 3D object detection regardless of input modalities.
However, potential drawbacks arise when aggregating both LiDAR features and camera features into a single object query, as the modalities from different domains may suffer from negative fusion when they are fused.
To prove that the proposed proximity-based modality ensemble module is effective to address this problem, we evaluate the MEFormer in LiDAR or camera corruption scenarios.
As presented in Table~\ref{sensor_corruption_table}, our framework shows the best performance in all scenarios.
Specifically, compared to MOAD, PME improves 0.9\% mAP in the beam reduction scenario and 0.3\% mAP in the camera dirt occlusion scenario.

Also, performance comparison in challenging environments is presented in Table~\ref{sensor_corruption_table}, and MEFormer outperforms other frameworks for all challenging environments.
Especially on rainy days where LiDAR shows noisy signals due to the refraction of the laser by raindrops, MEFormer shows a large performance gap (+1.2\% NDS and +1.7\% mAP) compared to CMT.
This result validates that our framework fully leverages the camera modality in scenarios where LiDAR struggles.
In addition, compared to MOAD, PME improves mAP by 0.6\% at night, where the camera modality shows noisy signals due to dark images.
This performance gap validates that PME adaptively exploits desirable modalities depending on the environment, avoiding negative fusion.
        \begin{table*}
\centering
\setlength{\tabcolsep}{14pt}
\begin{tabular}{lccccccc}
    \toprule
    \multirow{2}{*}{Method} &  \multirow{2}{*}{Modality} & \multicolumn{2}{c} {Near} & \multicolumn{2}{c} {Middle} & \multicolumn{2}{c}{Far} \\
    \cmidrule(lr){3-4} \cmidrule(lr){5-6} \cmidrule(lr){7-8}
    &  & NDS & mAP & NDS & mAP & NDS & mAP \\
    \midrule 
    \midrule
    PETR~\cite{liu2022petr}$\ddagger$ & C & 53.8 & 53.1 & 40.2 & 31.8 & 25.7 & 14.5 \\
    TransFusion-L~\cite{Bai2022transfusion}$\dagger$ & L  & 77.3 & 77.5 & 67.9 & 61.5 & 47.5 & 34.3 \\
    BEVFusion~\cite{liu2023bevfusion}$\dagger$ & L+C & 78.0 & 79.2 & 69.3 & 64.1 & 49.8 & 38.9 \\
    DeepInteraction~\cite{yang2022deepinteraction}$\ddagger$ & L+C & 75.3 & 78.6 & 67.9 & 65.4 & 48.4 & 40.8 \\
    \midrule
    CMT~\cite{yan2023cmt} & L+C & 79.4 & 81.0 & 70.9 & 65.8 & 52.6 & 42.5 \\
    \textbf{CMT w/ PME} & L+C & 79.6 & 81.1 & 71.1 & 66.1 & 52.9 & 42.9 \\
    \midrule
    \textbf{MOAD} & L+C & \textbf{80.4} & 82.3 & 71.7 & 66.8 & \textbf{53.4} & \textbf{43.6} \\
    \textbf{MOAD w/ PME} & L+C & \textbf{80.4} & \textbf{82.4} & \textbf{72.0} & \textbf{67.1} & 53.3 & \textbf{43.6} \\
    \bottomrule
\end{tabular}
\caption{
\textbf{Comparison of detection performance according to object distance.}
Near, middle, and far refer to distances under 20m, between 20m and 30m, and over 30m, respectively.
$\dagger$ and $\ddagger$ denote the results obtained using the OpenPCDet~\cite{openpcdet2020} reproduced weights and weight provided in the official GitHub repository respectively.
}
\label{distance_table}
\vspace{-12pt}
\end{table*}

\vspace{-12pt}
\subsubsection{Detection range}
\label{exp:distance}
\vspace{-6pt}
LiDAR sensor struggles to collect enough points of the objects that are located far away from the ego car.
This often leads to the failure to detect distant objects, which can be alleviated by utilizing the camera modality that has dense signals.
We show performance comparison across the various detection ranges to validate that MEFormer effectively utilizes the camera modality.
As shown in Table~\ref{distance_table}, MEFormer outperforms previous frameworks for distant objects.
Specifically, introducing MOAD achieves 1.1\% mAP improvement compared to CMT, which proves that MOAD helps extract geometric information from the camera modality.
In addition, PME improves the detection performance of objects located at middle and far distances compared to near distances.
This validates that PME addresses the negative fusion of noisy LiDAR information.
\vspace{-8pt}
    \section{Analysis}
\label{exp:ablations}
        \begin{table}[t!]
    \centering
    \setlength{\tabcolsep}{10pt}
    \begin{tabular}{c cc cc}
    \toprule
    \multicolumn{1}{c}{}
    &MOAD & PME & NDS & mAP \\
    \midrule
    \midrule
    (a)& & & 72.9 & 70.3 \\
    (b)& &\checkmark & 73.1 & 70.5 \\
    (c)&\checkmark & & 73.7 & 71.3 \\
    (d)&\checkmark  &\checkmark & \textbf{73.9} & \textbf{71.5} \\
    \bottomrule
    \end{tabular}
    \caption{
    \textbf{Ablations for proposed modules.}
    MOAD is modality-agnostic decoding and PME indicates proximity-based modality ensemble}
    \label{ablation_table}
    \vspace{-16pt}
\end{table}
\subsection{Ablation studies}
We present ablation studies of the proposed training strategy and module in Table~\ref{ablation_table}. 
All experiments are conducted on nuScenes validation set. 
First of all, as shown in (b), adapting PME improves detection performance by 0.2\% NDS and mAP compared to (a) which is identical to CMT.
Note that inputs $Z_L$ and $Z_C$ for the PME in (b) are generated by a transformer decoder trained with only the multi-modal decoding branch since our modality-agnostic decoding is not yet applied.
This shows that PME alone helps improve the detection performance.
Next, (c) shows applying modality-agnostic decoding enhances the performance by 0.8\% NDS and 1.0\% mAP.
Note that only a multi-modal decoding branch is used for the box prediction in (c) during inference time and there is no additional computation compared to (a).
This empirical evidence verifies that reducing reliance on a LiDAR sensor improves the detection performance in overall environments.
In (d), we extend our methodology to include both modality-agnostic decoding and proximity-based modality ensemble.
Applying both MOAD and PME shows 1.0\% NDS and 1.2\% mAP performance gain compared to (a), resulting in state-of-the-art detection performance.
Performance comparison between (c) and (d) in challenging environments is discussed in Section~\ref{exp:sensor_corruption} and shows a larger performance gap compared to that in the overall environments, which means PME is more effective as the environment becomes more challenging.
        \begin{table}[t!]
    \centering
    \setlength{\tabcolsep}{9pt}
    \begin{tabular}{l cc c}
    \toprule
    Method & FPS & NDS & mAP \\
    \midrule
    \midrule
    DeepInteractioion     & 1.5 & 72.6 & 69.9 \\
    SparseFusion          & 2.3 & 72.8 & 70.4 \\
    CMT                   & \textbf{3.4} & 72.9 & 70.3 \\
    \textbf{Ours}         & 3.1 & \textbf{73.9} & \textbf{71.5} \\
    \bottomrule
    \end{tabular}
    \caption{
    \textbf{Comparison of inference speed and detection performance.}
    Voxelization time is included in the inference time following CMT~\cite{yan2023cmt}.}
    \label{speed_table}
    \vspace{-16pt}
\end{table}
\subsection{Inference speed}
We compare the inference speed of our framework with previous frameworks and the result is shown in Table~\ref{speed_table}.
\begin{figure*}[t!]
    \centering
    \includegraphics[width=\linewidth]{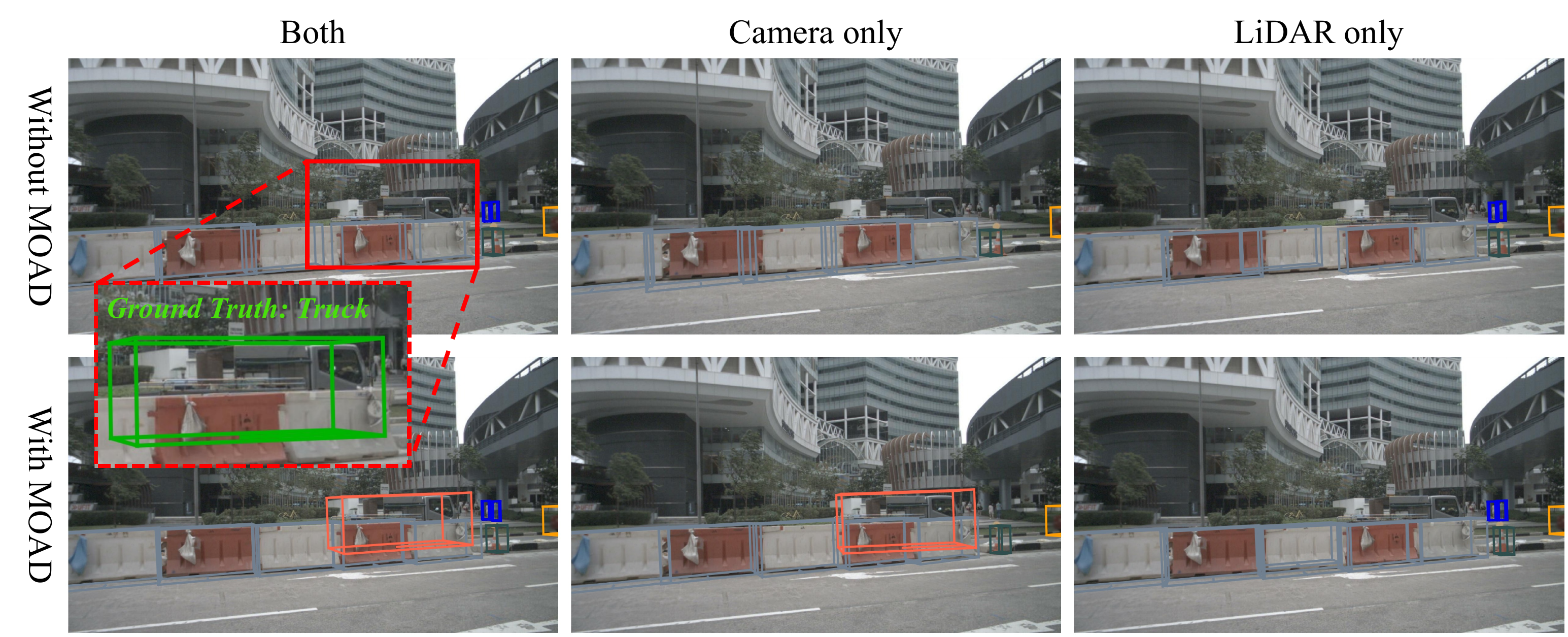}
    \vspace{-17pt}
    \caption{
        \textbf{Qualitative results of MOAD in nuScenes validation set.}
        With MOAD detects a truck with the help of geometric information in the camera modality while without MOAD fails.
    }
    \label{fig:moad_analysis}
\end{figure*}
\begin{figure*}[ht]
    \centering
    \includegraphics[width=0.95\textwidth]{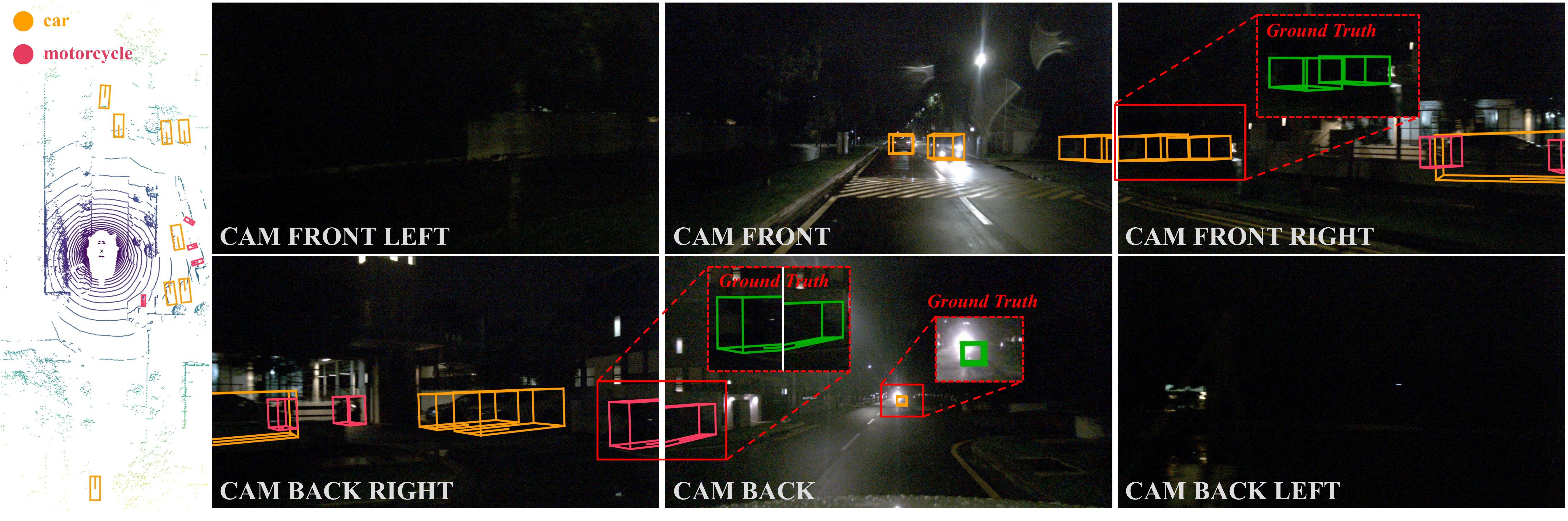}
    \vspace{-6pt}
    \caption{
        \textbf{Qualitative results on multi-view images and BEV space at night time in nuScenes validation set.}
        MEFormer shows promising detection results for objects that are difficult to identify with the cameras.
        We provide additional ground truth boxes for those objects to recognize easily in images.
    }
    \label{fig:qualitative_figure}
    \vspace{-14pt}
\end{figure*}
MEFormer shows 1.0\% NDS and 1.2\% mAP performance improvement with only a 0.3 FPS reduction in inference speed.
Note that all results are measured on a single NVIDIA RTX 3090 GPU and voxelization time is included following~\cite{yan2023cmt}.
    \subsection{Qualitative results}
\label{exp:qualitative}
In this section, we validate the effectiveness of our framework with qualitative results.
First, Figure~\ref{fig:moad_analysis} shows that the transformer decoder trained with MOAD utilizes the camera to successfully localize the truck that LiDAR fails to detect, resulting in bounding boxes in the multi-modal decoding branch as well.
However, without MOAD, all decoding branches fail to detect the same truck.
This result validates that MOAD reduces the LiDAR reliance problem in the modality fusion process and helps the framework utilize geometric information in camera modality.
\vspace{-3pt}

Figure~\ref{fig:qualitative_figure} shows qualitative results of MEFormer in the dark environment, where the cameras show extremely corrupted signals.
In the front right camera, our framework successfully detects two cars that are hard to recognize with the camera.
Additionally, in the back view camera, our framework also detected the car in which a camera shows corrupted signals due to the car's headlight.
Qualitative results validate that our framework shows competitive detection performance in challenging environments.
\section{Conclusion}
\label{sec:conclusion}
In this paper, we present MEFormer, an effective framework to fully leverage the LiDAR sensor and camera sensors, addressing the LiDAR reliance problem.
MEFormer is trained to extract both geometric and semantic information from each modality using the modality-agnostic decoding training strategy, resulting in promising results in various LiDAR malfunction scenarios as well as overall environments.
In addition, the proximity-based modality ensemble module shows another performance improvement by preventing negative fusion in challenging environments.
Extensive experiments validate that MEFormer achieves state-of-the-art performance in various scenarios.
We hope that MEFormer can inspire further research in the field of robust multi-modal 3D object detection.

\clearpage
{
    \small
    \bibliographystyle{ieeenat_fullname}
    \bibliography{main}

\begin{thebibliography}{47}
\providecommand{\natexlab}[1]{#1}
\providecommand{\url}[1]{\texttt{#1}}
\expandafter\ifx\csname urlstyle\endcsname\relax
  \providecommand{\doi}[1]{doi: #1}\else
  \providecommand{\doi}{doi: \begingroup \urlstyle{rm}\Url}\fi

\bibitem[Bai et~al.(2022)Bai, Hu, Zhu, Huang, Chen, Fu, and Tai]{Bai2022transfusion}
Xuyang Bai, Zeyu Hu, Xinge Zhu, Qingqiu Huang, Yilun Chen, Hongbo Fu, and Chiew{-}Lan Tai.
\newblock Transfusion: Robust lidar-camera fusion for 3d object detection with transformers.
\newblock In \emph{CVPR}, 2022.

\bibitem[Caesar et~al.(2020)Caesar, Bankiti, Lang, Vora, Liong, Xu, Krishnan, Pan, Baldan, and Beijbom]{caesar2020nuscenes}
Holger Caesar, Varun Bankiti, Alex~H Lang, Sourabh Vora, Venice~Erin Liong, Qiang Xu, Anush Krishnan, Yu Pan, Giancarlo Baldan, and Oscar Beijbom.
\newblock nuscenes: A multimodal dataset for autonomous driving.
\newblock In \emph{CVPR}, 2020.

\bibitem[Chen et~al.(2023)Chen, Zhang, Wang, Wang, and Zhao]{chen2023futr3d}
Xuanyao Chen, Tianyuan Zhang, Yue Wang, Yilun Wang, and Hang Zhao.
\newblock Futr3d: A unified sensor fusion framework for 3d detection.
\newblock In \emph{CVPR}, 2023.

\bibitem[Chen et~al.(2022)Chen, Li, Zhang, Fang, Jiang, and Zhao]{chen2022autoalignv2}
Zehui Chen, Zhenyu Li, Shiquan Zhang, Liangji Fang, Qinhong Jiang, and Feng Zhao.
\newblock Autoalignv2: Deformable feature aggregation for dynamic multi-modal 3d object detection.
\newblock In \emph{ECCV}, 2022.

\bibitem[Contributors(2020)]{mmdet3d2020}
MMDetection3D Contributors.
\newblock {MMDetection3D: OpenMMLab} next-generation platform for general {3D} object detection.
\newblock \url{https://github.com/open-mmlab/mmdetection3d}, 2020.

\bibitem[Ge et~al.(2023)Ge, Chen, Xie, Wang, Hong, Lu, Li, and Luo]{ge2023metabev}
Chongjian Ge, Junsong Chen, Enze Xie, Zhongdao Wang, Lanqing Hong, Huchuan Lu, Zhenguo Li, and Ping Luo.
\newblock Metabev: Solving sensor failures for 3d detection and map segmentation.
\newblock In \emph{ICCV}, 2023.

\bibitem[Huang et~al.(2021)Huang, Huang, Zhu, Yun, and Du]{huang2021bevdet}
Junjie Huang, Guan Huang, Zheng Zhu, Ye Yun, and Dalong Du.
\newblock Bevdet: High-performance multi-camera 3d object detection in bird-eye-view.
\newblock \emph{arXiv preprint arXiv:2112.11790}, 2021.

\bibitem[Lee and Park(2020)]{lee2020vovnet}
Youngwan Lee and Jongyoul Park.
\newblock Centermask: Real-time anchor-free instance segmentation.
\newblock In \emph{CVPR}, 2020.

\bibitem[Li et~al.(2022{\natexlab{a}})Li, Chen, Qi, Li, Sun, and Jia]{li2022uvtr}
Yanwei Li, Yilun Chen, Xiaojuan Qi, Zeming Li, Jian Sun, and Jiaya Jia.
\newblock Unifying voxel-based representation with transformer for 3d object detection.
\newblock In \emph{NeurIPS}, 2022{\natexlab{a}}.

\bibitem[Li et~al.(2022{\natexlab{b}})Li, Yu, Meng, Caine, Ngiam, Peng, Shen, Lu, Zhou, Le, Yuille, and Tan]{li2022deepfusion}
Yingwei Li, Adams~Wei Yu, Tianjian Meng, Ben Caine, Jiquan Ngiam, Daiyi Peng, Junyang Shen, Yifeng Lu, Denny Zhou, Quoc~V. Le, Alan Yuille, and Mingxing Tan.
\newblock Deepfusion: Lidar-camera deep fusion for multi-modal 3d object detection.
\newblock In \emph{CVPR}, 2022{\natexlab{b}}.

\bibitem[Li et~al.(2023)Li, Ge, Yu, Yang, Wang, Shi, Sun, and Li]{li23bevdepth}
Yinhao Li, Zheng Ge, Guanyi Yu, Jinrong Yang, Zengran Wang, Yukang Shi, Jianjian Sun, and Zeming Li.
\newblock Bevdepth: Acquisition of reliable depth for multi-view 3d object detection.
\newblock In \emph{AAAI}, 2023.

\bibitem[Li et~al.(2022{\natexlab{c}})Li, Wang, Li, Xie, Sima, Lu, Qiao, and Dai]{li2022bevformer}
Zhiqi Li, Wenhai Wang, Hongyang Li, Enze Xie, Chonghao Sima, Tong Lu, Yu Qiao, and Jifeng Dai.
\newblock Bevformer: Learning bird's-eye-view representation from multi-camera images via spatiotemporal transformers.
\newblock In \emph{ECCV}, 2022{\natexlab{c}}.

\bibitem[Liang et~al.(2022)Liang, Xie, Yu, Xia, Lin, Wang, Tang, Wang, and Tang]{Liang2022bevfusion}
Tingting Liang, Hongwei Xie, Kaicheng Yu, Zhongyu Xia, Zhiwei Lin, Yongtao Wang, Tao Tang, Bing Wang, and Zhi Tang.
\newblock Bevfusion: {A} simple and robust lidar-camera fusion framework.
\newblock In \emph{NeurIPS}, 2022.

\bibitem[Liu et~al.(2022)Liu, Wang, Zhang, and Sun]{liu2022petr}
Yingfei Liu, Tiancai Wang, Xiangyu Zhang, and Jian Sun.
\newblock Petr: Position embedding transformation for multi-view 3d object detection.
\newblock In \emph{ECCV}, 2022.

\bibitem[Liu et~al.(2023{\natexlab{a}})Liu, Yan, Jia, Li, Gao, Wang, and Zhang]{liu2023petrv2}
Yingfei Liu, Junjie Yan, Fan Jia, Shuailin Li, Aqi Gao, Tiancai Wang, and Xiangyu Zhang.
\newblock Petrv2: A unified framework for 3d perception from multi-camera images.
\newblock In \emph{ICCV}, 2023{\natexlab{a}}.

\bibitem[Liu et~al.(2023{\natexlab{b}})Liu, Tang, Amini, Yang, Mao, Rus, and Han]{liu2023bevfusion}
Zhijian Liu, Haotian Tang, Alexander Amini, Xinyu Yang, Huizi Mao, Daniela~L Rus, and Song Han.
\newblock Bevfusion: Multi-task multi-sensor fusion with unified bird's-eye view representation.
\newblock In \emph{ICRA}, 2023{\natexlab{b}}.

\bibitem[Loshchilov and Hutter(2017)]{loshcilov2019cosine}
Ilya Loshchilov and Frank Hutter.
\newblock {SGDR:} stochastic gradient descent with warm restarts.
\newblock In \emph{ICLR}, 2017.

\bibitem[Loshchilov and Hutter(2019)]{loshchilov2019adamw}
Ilya Loshchilov and Frank Hutter.
\newblock Decoupled weight decay regularization.
\newblock In \emph{ICLR}, 2019.

\bibitem[Mao et~al.(2023)Mao, Shi, Wang, and Li]{mao2023survey}
Jiageng Mao, Shaoshuai Shi, Xiaogang Wang, and Hongsheng Li.
\newblock 3d object detection for autonomous driving: {A} comprehensive survey.
\newblock \emph{IJCV}, 2023.

\bibitem[Nekrasov et~al.(2019)Nekrasov, Dharmasiri, Spek, Drummond, Shen, and Reid]{nekrasov2019mtl}
Vladimir Nekrasov, Thanuja Dharmasiri, Andrew Spek, Tom Drummond, Chunhua Shen, and Ian~D. Reid.
\newblock Real-time joint semantic segmentation and depth estimation using asymmetric annotations.
\newblock In \emph{ICRA}, 2019.

\bibitem[Philion and Fidler(2020)]{philion2020lift}
Jonah Philion and Sanja Fidler.
\newblock Lift, splat, shoot: Encoding images from arbitrary camera rigs by implicitly unprojecting to 3d.
\newblock In \emph{ECCV}, 2020.

\bibitem[Qi et~al.(2018)Qi, Liu, Wu, Su, and Guibas]{qi2018frustum}
Charles~R Qi, Wei Liu, Chenxia Wu, Hao Su, and Leonidas~J Guibas.
\newblock Frustum pointnets for 3d object detection from rgb-d data.
\newblock In \emph{CVPR}, 2018.

\bibitem[Reading et~al.(2021)Reading, Harakeh, Chae, and Waslander]{reading2021categorical}
Cody Reading, Ali Harakeh, Julia Chae, and Steven~L Waslander.
\newblock Categorical depth distribution network for monocular 3d object detection.
\newblock In \emph{CVPR}, 2021.

\bibitem[Smith(2017)]{smith2017cyclic}
Leslie~N. Smith.
\newblock Cyclical learning rates for training neural networks.
\newblock In \emph{WACV}, 2017.

\bibitem[Song et~al.(2024)Song, Liu, Jia, Luo, Zhang, Yang, Wang, and Jia]{song2024robustness}
Ziying Song, Lin Liu, Feiyang Jia, Yadan Luo, Guoxin Zhang, Lei Yang, Li Wang, and Caiyan Jia.
\newblock Robustness-aware 3d object detection in autonomous driving: A review and outlook.
\newblock \emph{arXiv preprint arXiv:2401.06542}, 2024.

\bibitem[Team(2020)]{openpcdet2020}
OpenPCDet~Development Team.
\newblock Openpcdet: An open-source toolbox for 3d object detection from point clouds.
\newblock \url{https://github.com/open-mmlab/OpenPCDet}, 2020.

\bibitem[Vora et~al.(2020)Vora, Lang, Helou, and Beijbom]{vora2020pointpainting}
Sourabh Vora, Alex~H Lang, Bassam Helou, and Oscar Beijbom.
\newblock Pointpainting: Sequential fusion for 3d object detection.
\newblock In \emph{CVPR}, 2020.

\bibitem[Wang et~al.(2021{\natexlab{a}})Wang, Ma, Zhu, and Yang]{wang2021pointaugmenting}
Chunwei Wang, Chao Ma, Ming Zhu, and Xiaokang Yang.
\newblock Pointaugmenting: Cross-modal augmentation for 3d object detection.
\newblock In \emph{CVPR}, 2021{\natexlab{a}}.

\bibitem[Wang et~al.(2023{\natexlab{a}})Wang, Tang, Shi, Li, Li, Schiele, and Wang]{wang2023unitr}
Haiyang Wang, Hao Tang, Shaoshuai Shi, Aoxue Li, Zhenguo Li, Bernt Schiele, and Liwei Wang.
\newblock Unitr: A unified and efficient multi-modal transformer for bird's-eye-view representation.
\newblock In \emph{ICCV}, 2023{\natexlab{a}}.

\bibitem[Wang et~al.(2023{\natexlab{b}})Wang, Zhou, Li, and Ren]{wang2023challenge}
Ke Wang, Tianqiang Zhou, Xingcan Li, and Fan Ren.
\newblock Performance and challenges of 3d object detection methods in complex scenes for autonomous driving.
\newblock \emph{{IEEE} Trans. Intell. Veh.}, 2023{\natexlab{b}}.

\bibitem[Wang et~al.(2021{\natexlab{b}})Wang, Zhu, Pang, and Lin]{wang2021fcos3d}
Tai Wang, Xinge Zhu, Jiangmiao Pang, and Dahua Lin.
\newblock Fcos3d: Fully convolutional one-stage monocular 3d object detection.
\newblock In \emph{ICCV}, 2021{\natexlab{b}}.

\bibitem[Wang et~al.(2019)Wang, Chao, Garg, Hariharan, Campbell, and Weinberger]{wang2019pseudo}
Yan Wang, Wei-Lun Chao, Divyansh Garg, Bharath Hariharan, Mark Campbell, and Kilian~Q Weinberger.
\newblock Pseudo-lidar from visual depth estimation: Bridging the gap in 3d object detection for autonomous driving.
\newblock In \emph{CVPR}, 2019.

\bibitem[Wang et~al.(2021{\natexlab{c}})Wang, Guizilini, Zhang, Wang, Zhao, and Solomon]{wang2021detr3d}
Yue Wang, Vitor Guizilini, Tianyuan Zhang, Yilun Wang, Hang Zhao, and Justin Solomon.
\newblock {DETR3D:} 3d object detection from multi-view images via 3d-to-2d queries.
\newblock In \emph{CoRL}, 2021{\natexlab{c}}.

\bibitem[Xie et~al.(2023{\natexlab{a}})Xie, Kong, Zhang, Ren, Pan, Chen, and Liu]{xie2023robobev}
Shaoyuan Xie, Lingdong Kong, Wenwei Zhang, Jiawei Ren, Liang Pan, Kai Chen, and Ziwei Liu.
\newblock Robobev: Towards robust bird's eye view perception under corruptions.
\newblock \emph{arXiv preprint arXiv:2304.06719}, 2023{\natexlab{a}}.

\bibitem[Xie et~al.(2023{\natexlab{b}})Xie, Xu, Rakotosaona, Rim, Tombari, Keutzer, Tomizuka, and Zhan]{xie2023sparsefusion}
Yichen Xie, Chenfeng Xu, Marie-Julie Rakotosaona, Patrick Rim, Federico Tombari, Kurt Keutzer, Masayoshi Tomizuka, and Wei Zhan.
\newblock Sparsefusion: Fusing multi-modal sparse representations for multi-sensor 3d object detection.
\newblock In \emph{ICCV}, 2023{\natexlab{b}}.

\bibitem[Yan et~al.(2023)Yan, Liu, Sun, Jia, Li, Wang, and Zhang]{yan2023cmt}
Junjie Yan, Yingfei Liu, Jianjian Sun, Fan Jia, Shuailin Li, Tiancai Wang, and Xiangyu Zhang.
\newblock Cross modal transformer: Towards fast and robust 3d object detection.
\newblock In \emph{ICCV}, 2023.

\bibitem[Yan et~al.(2018)Yan, Mao, and Li]{yan2018second}
Yan Yan, Yuxing Mao, and Bo Li.
\newblock Second: Sparsely embedded convolutional detection.
\newblock \emph{Sensors}, 2018.

\bibitem[Yang et~al.(2023)Yang, Chen, Tian, Tao, Zhu, Zhang, Huang, Li, Qiao, Lu, et~al.]{yang2023bevformerv2}
Chenyu Yang, Yuntao Chen, Hao Tian, Chenxin Tao, Xizhou Zhu, Zhaoxiang Zhang, Gao Huang, Hongyang Li, Yu Qiao, Lewei Lu, et~al.
\newblock Bevformer v2: Adapting modern image backbones to bird's-eye-view recognition via perspective supervision.
\newblock In \emph{CVPR}, 2023.

\bibitem[Yang et~al.(2022)Yang, Chen, Miao, Li, Zhu, and Zhang]{yang2022deepinteraction}
Zeyu Yang, Jiaqi Chen, Zhenwei Miao, Wei Li, Xiatian Zhu, and Li Zhang.
\newblock Deepinteraction: 3d object detection via modality interaction.
\newblock In \emph{NeurIPS}, 2022.

\bibitem[Yin et~al.(2021{\natexlab{a}})Yin, Zhou, and Kr{\"a}henb{\"u}hl]{yin2021center}
Tianwei Yin, Xingyi Zhou, and Philipp Kr{\"a}henb{\"u}hl.
\newblock Center-based 3d object detection and tracking.
\newblock In \emph{CVPR}, 2021{\natexlab{a}}.

\bibitem[Yin et~al.(2021{\natexlab{b}})Yin, Zhou, and Kr{\"{a}}henb{\"{u}}hl]{yin2021mvp}
Tianwei Yin, Xingyi Zhou, and Philipp Kr{\"{a}}henb{\"{u}}hl.
\newblock Multimodal virtual point 3d detection.
\newblock In \emph{NeurIPS}, 2021{\natexlab{b}}.

\bibitem[Yu et~al.(2023)Yu, Tao, Xie, Lin, Liang, Wang, Chen, Hao, Wang, and Liang]{yu2023benchmark}
Kaicheng Yu, Tang Tao, Hongwei Xie, Zhiwei Lin, Tingting Liang, Bing Wang, Peng Chen, Dayang Hao, Yongtao Wang, and Xiaodan Liang.
\newblock Benchmarking the robustness of lidar-camera fusion for 3d object detection.
\newblock In \emph{CVPRW}, 2023.

\bibitem[Zhang et~al.(2019)Zhang, Zhou, Sun, Wang, Shi, and Loy]{zhang2019robust}
Wenwei Zhang, Hui Zhou, Shuyang Sun, Zhe Wang, Jianping Shi, and Chen~Change Loy.
\newblock Robust multi-modality multi-object tracking.
\newblock In \emph{ICCV}, 2019.

\bibitem[Zhang and Yang(2022)]{zhang2022mtl_survey}
Yu Zhang and Qiang Yang.
\newblock A survey on multi-task learning.
\newblock \emph{{IEEE} Trans. Knowl. Data Eng.}, 2022.

\bibitem[Zhou and Tuzel(2018)]{zhou2018voxelnet}
Yin Zhou and Oncel Tuzel.
\newblock Voxelnet: End-to-end learning for point cloud based 3d object detection.
\newblock In \emph{CVPR}, 2018.

\bibitem[Zhu et~al.(2019)Zhu, Jiang, Zhou, Li, and Yu]{zhu2019cbgs}
Benjin Zhu, Zhengkai Jiang, Xiangxin Zhou, Zeming Li, and Gang Yu.
\newblock Class-balanced grouping and sampling for point cloud 3d object detection.
\newblock \emph{arXiv preprint arXiv:1908.09492}, 2019.

\bibitem[Zhu et~al.(2023)Zhu, Zhang, Chen, Dong, Zhao, Ding, Zhong, and Zheng]{zhu2023understanding}
Zijian Zhu, Yichi Zhang, Hai Chen, Yinpeng Dong, Shu Zhao, Wenbo Ding, Jiachen Zhong, and Shibao Zheng.
\newblock Understanding the robustness of 3d object detection with bird's-eye-view representations in autonomous driving.
\newblock In \emph{CVPR}, 2023.

\end{thebibliography}
}

% WARNING: do not forget to delete the supplementary pages from your submission 
% \maketitlesupplementary
% \input{supple/X_suppl}

\end{document}